\documentclass[letterpaper, 10 pt, conference]{ieeeconf}  

\IEEEoverridecommandlockouts                              

\usepackage{graphicx} 
\usepackage{mathptmx} 
\usepackage{amsmath} 
\usepackage{amssymb}  
\usepackage{comment}

\usepackage{subcaption}
\usepackage{booktabs}
\usepackage{multirow}
\usepackage{xurl}
\usepackage{array}
\usepackage{arydshln}
\usepackage{eso-pic}

\title{\LARGE \bf
A Multimodal Data-Collection Framework\\Linking Avatar-Centric Observations and Operator-Centric Signals\\Through a Whole-Body Android Avatar
}

\author{Kaoruko Shinkawa$^{1}$, Yuma Kabe$^{1}$, Chenfei Xu$^{2}$, Yuya Okadome$^{3}$, Yutaka Nakamura$^{2}$, and Yoshihiro Nakata$^{1}$
\thanks{*This work was supported by JST Moonshot R\&D Grant Number JPMJMS2011. One of the authors is supported by JST SPRING, Grant Number JPMJSP2131 (Scholarship award).}
\thanks{$^{1}$Kaoruko Shinkawa, Yuma Kabe and Yoshihiro Nakata are with The University of Electro-Communications, Tokyo, Japan
        {\tt\small \{kshinkawa, yuma.kabe, ynakata\}@uec.ac.jp}}%
\thanks{$^{2}$Chenfei Xu and Yutaka Nakamura are with GRP, RIKEN, Kyoto, Japan
        {\tt\small \{chenfei.xu, yutaka.nakamura\}@riken.jp}}%
\thanks{$^{3}$Yuya Okadome is with Tokyo University of Science, Tokyo, Japan
        {\tt\small okadome@rs.tus.ac.jp}}%
}

\begin{document}

\maketitle
\thispagestyle{empty}
\pagestyle{empty}
\AddToShipoutPictureFG*{%
  \AtPageLowerLeft{%
    \put(\LenToUnit{0.75in},\LenToUnit{0.30in}){%
      \parbox[b]{7in}{%
        \normalfont\fontsize{8}{9}\selectfont
        \raggedright
        \copyright~2026 IEEE.
        Personal use of this material is permitted.
        Permission from IEEE must be obtained for all
        other uses, in any current or future media,
        including reprinting/republishing this material
        for advertising or promotional purposes,
        creating new collective works, for resale or
        redistribution to servers or lists, or reuse of
        any copyrighted component of this work in
        other works.
      }%
    }%
  }%
}

\begin{abstract}
Robots interacting with humans must adapt their behavior not only to the surrounding environment but also to the actions and responses of humans.
Because human behavioral responses depend on the social context and can influence subsequent interactions, data for developing socially interactive robots should capture both robot-observable situations and the corresponding human behavioral responses.
This study developed a multimodal data-collection framework that uses a teleoperated whole-body android avatar.
The framework records avatar-centric observations (ACOs), including visual and auditory observations, received control commands, and avatar motor states, together with operator-centric signals (OCSs), including gaze, facial expressions, head and body motion, and operator-derived control information.
The recordings contain device timestamps for offline temporal association, and operator gaze can be spatially associated with avatar-side visual observations.
We deployed the framework in two distinct real-world social settings, a university open-lab event and a corporate office, and collected 4.5 hours of multimodal recordings, from which 2.6 hours of valid social-interaction segments were extracted.
Exploratory analyses of operator gaze, facial expressions, and speech illustrated how linked ACOs and OCSs can be used to examine operator social cues in relation to avatar-observable interaction contexts, providing a foundation for future social-interaction models.

\end{abstract}

\section{Introduction}
\label{sec:introduction}

Robots operating alongside humans must adapt not only to physical states but also to human speech and nonverbal behavior~\cite{Sirithunge2019}.
While learning for physical tasks often involves modeling relationships between environmental states and robot actions~\cite{Kawaharazuka2024,He2025}, social interaction requires accounting for the dependence of human responses on preceding exchanges and social context~\cite{Fong2003}.
Our long-term goal is to develop a social-interaction world model that captures the progression of such interactions over time~\cite{Nakamura2026Workshop} and ultimately supports robot autonomy by enabling robots to consider how their actions may affect subsequent human responses and the course of an interaction.
This study addresses the multimodal data collection prerequisite for pursuing that goal.

The development of such models requires data that describe both interaction contexts observable from an embodied robot or avatar and human responses produced within these contexts.
External cameras record the overall interaction scenario ~\cite{Fieraru2020,Joo2019}, whereas sensors and controllers on a robot or avatar provide information from the embodiment itself, including egocentric visual and auditory observations~\cite{Zhang2022,NorthCutt2020}, commands received by the avatar, and avatar motor states.
These data are collectively referred to as avatar-centric observations (ACOs).
In parallel, we define operator gaze, facial expressions, head and body motion, and operator-generated control inputs as operator-centric signals (OCSs).
These signals provide observable measures related to attention allocation and behavioral responses, rather than direct measurements of internal cognitive states or intentions.
Thus, recording and associating ACOs and OCSs provides data for learning human behavioral responses to interaction contexts observed through the avatar, laying the foundation for developing a social-interaction world model that could potentially be adapted to different robot embodiments.

A teleoperated whole-body android avatar provides both perspectives within a single embodied interaction.
The operator perceives and interacts in the remote environment through the avatar while generating gaze, facial expressions, body motions, and control signals, which are embodied through the avatar during the same interaction.
Building on the android avatar ``Yui'' and its immersive teleoperation interface~\cite{Shinkawa2026Yui}, we developed and implemented a multimodal data-collection framework that records and links ACOs and OCSs generated from the same avatar-mediated social interaction.

We deployed the proposed framework in a university open-lab setting and a corporate office, collecting 4.5 hours of multimodal recordings, from which 2.6 hours of valid social-interaction segments were extracted.
As an exploratory use of the linked recordings, we examined face-directed gaze and smile onset in relation to subsequent speech onset, as well as the occurrence of operator social cues across speech states.
These analyses illustrate multiple temporal and contextual uses of the collected data rather than establish generalizable behavioral findings.

The contributions of this study are as follows:
\begin{itemize}
    \item Development of a multimodal data-collection framework that records and associates ACOs and OCSs generated from the same embodied social interaction using a teleoperated whole-body android avatar.
    \item Deployment of the framework in two distinct real-world environments, yielding 4.5 hours of multimodal recordings and 2.6 hours of valid social-interaction segments.
    \item Temporal and contextual analyses
    of operator social cues using the linked ACOs and OCSs.
\end{itemize}

\section{Related Work}
\label{sec:related_work}
\begin{figure}[t]
    \centering
    \includegraphics[width=0.85\linewidth]{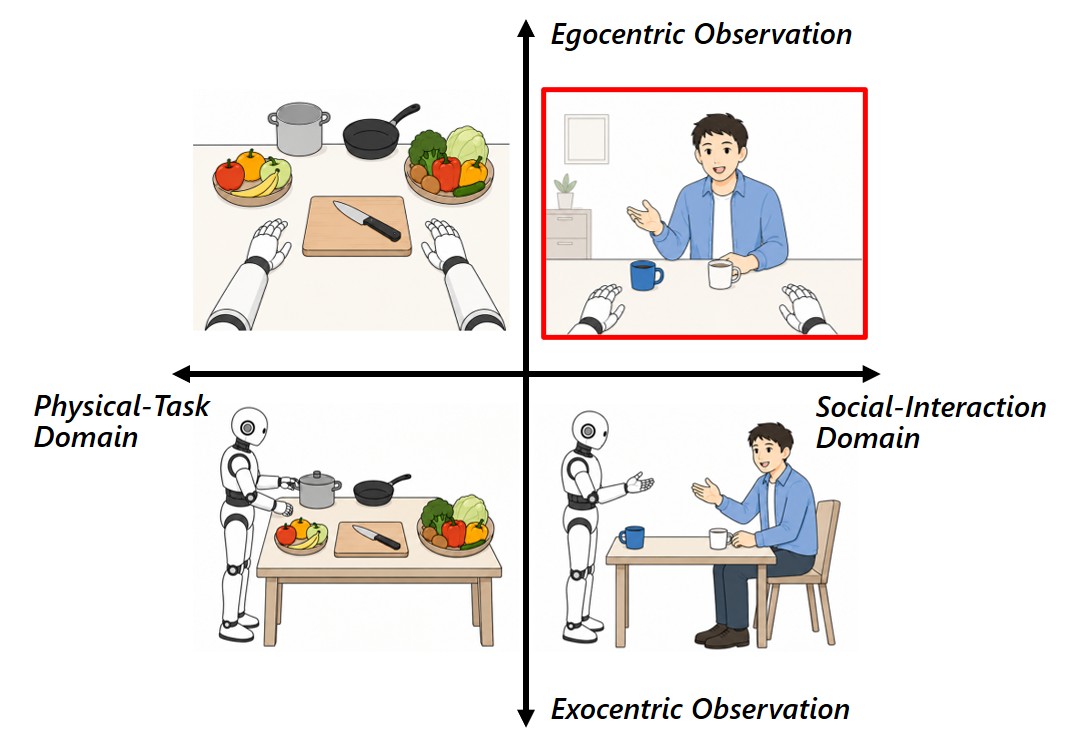}
    \caption{Conceptual positioning of this study along the task-domain and observation-perspective axes. The proposed framework lies in the egocentric social-interaction quadrant.}
    \label{fig:concept_map}
\end{figure}

Fig.~\ref{fig:concept_map} illustrates the positioning of the present study along two axes---task domain and observation perspective---that organize the discussion below.

\subsection{Task Domain: From Physical Tasks to Social Interaction}

The horizontal axis contrasts physical tasks with social interaction.
Robot learning and world-model research have largely focused on manipulation, navigation, and locomotion, modeling how robot actions affect objects and environments~\cite{Kawaharazuka2024,Sakagami2023,Wu2023}.
Recent work has also leveraged real-world egocentric and teleoperation data to learn behaviors transferable across embodiments and environments~\cite{He2025}.
However, in social interaction, human responses are shaped by robot appearance, dialogue, gaze, and motion~\cite{Fong2003,Admoni2017,Dragan2013}.
Toward autonomous social interaction, data should capture robot-observable social context and interlocutor behavior alongside corresponding human-demonstrated social responses within the same interaction.


\subsection{Observation Perspectives: Exocentric vs. Egocentric}

The vertical axis contrasts exocentric and egocentric observation.
Exocentric approaches capture the overall interaction scene, including posture, distance, speech, and actions~\cite{Fieraru2020,Joo2019}, whereas egocentric approaches provide first-person observations from the viewpoint of a participant or embodied agent~\cite{Zhang2022,NorthCutt2020}.
Compared with participant-worn sensing, teleoperation allows an operator to perceive the interaction from the avatar's viewpoint and generate social responses through the same robotic embodiment, linking human-demonstrated responses to information available to the robot.

RoboTrio2~\cite{EliseiHB24} is a closely related multimodal corpus that links robot-side sensory streams with human pilot behavior during immersive teleoperation. 
It synchronously records the robot's stereo views and binaural audio together with the pilot's gaze, head, jaw/lip movements, and robot motion streams. 
For gaze analysis, the pilot's gaze is categorized into four predefined targets: the left user, right user, tablet, and elsewhere.
Table~\ref{tab:related_frameworks} compares representative egocentric social-interaction data-collection approaches in terms of recorded modalities, view--operator pairing, and scene linkage.



\begin{table}[t]
\caption{Comparison of egocentric social-interaction
data-collection approaches.}
\label{tab:related_frameworks}
\centering
\scriptsize
\setlength{\tabcolsep}{2pt}
\renewcommand{\arraystretch}{1.08}
\begin{tabular}{@{}
>{\raggedright\arraybackslash}p{0.18\columnwidth}
>{\raggedright\arraybackslash}p{0.43\columnwidth}
>{\centering\arraybackslash}p{0.13\columnwidth}
>{\raggedright\arraybackslash}p{0.20\columnwidth}
@{}}
\hline
Work &
Recorded data &
VOP$^{\dagger}$ &
Scene linkage \\
\hline
EgoBody~\cite{Zhang2022} &
Wearer RGB-D, gaze, head, and hands &
No &
Image-plane gaze \\[3pt]

EgoCom~\cite{NorthCutt2020} &
Participant-worn video and stereo audio &
No &
No gaze signal \\[3pt]

RoboTrio2~\cite{EliseiHB24} &
Robot stereo views and audio; pilot gaze, head, jaw/lips,
and motion &
Yes &
Predefined gaze targets \\[3pt]

\textbf{This study} &
\textbf{Avatar visual, audio, control, and motor data;
operator gaze, face, head, and upper-body signals} &
\textbf{Yes} &
\textbf{Image-mapped gaze; derivable acoustic map} \\[3pt]
\hline
\end{tabular}

\vspace{1mm}
\scriptsize{
$^{\dagger}$VOP: View--operator pairing (association between a viewpoint and its operator).
}
\end{table}

Our previous workshop paper introduced the concept of a social-interaction world model and identified linked multimodal data collection as a prerequisite for its development~\cite{Nakamura2026Workshop}.
This study implements that concept using a whole-body android avatar.
RoboTrio2 and the framework developed in this study both link robot-side and teleoperator data, but differ in the scope of recorded signals and their linkage to the interaction scene.
For example, image-mapped gaze and facial blendshape signals supported the face-directed-gaze and smile analyses, respectively, in Section~\ref{sec:analysis}.
Together with control commands and avatar motor states, these signals enable operator responses to be examined in relation to avatar-observable contexts and avatar behavior within the same interaction, providing data relevant to future autonomous social interaction.


\section{Implementation of the Multimodal Data-Collection Framework}
\label{sec:framework}

\begin{figure}[t]
    \centering
    \includegraphics[width=\linewidth]{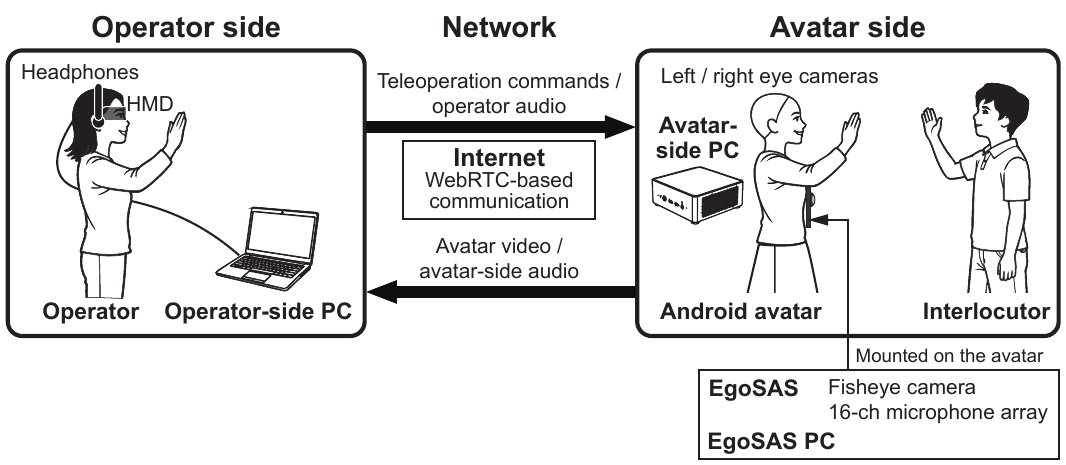}
    \caption{Overview of the teleoperated android avatar system used for multimodal data-collection. 
    EgoSAS denotes the ego-centric situational awareness sensor.}
    \label{fig:teleop_system}
\end{figure}

The proposed framework uses a teleoperated whole-body android avatar to record ACOs and OCSs from the same embodied social interaction.
This section describes the avatar,  teleoperation system, recorded modalities, and their temporal and spatial association.
Fig.~\ref{fig:teleop_system} shows the overall system configuration, and Table~\ref{tab:recorded_modalities} summarizes the recorded modalities.


\begin{table*}[t]
\centering
\caption{Modalities recorded as avatar-centric observations (ACOs) and operator-centric signals (OCSs) by the proposed framework.}
\label{tab:recorded_modalities}
\footnotesize
\setlength{\tabcolsep}{4pt}
\begin{tabular}{@{}llllll@{}}
\hline
Category & Recorded modality & Recording device & Source & Rate & Format \\
\hline

ACO
& Fisheye images
& EgoSAS PC
& EgoSAS fisheye camera
& 30 fps
& JPEG \\

ACO
& 16-channel audio
& EgoSAS PC
& EgoSAS microphone array
& 44.1 kHz
& WAV \\

ACO
& Eye-camera images
& Operator-side PC
& Avatar eye camera
& 10 fps
& JPEG \\

ACO
& Received control commands
& Avatar-side PC
& Avatar-side controller
& 10 Hz
& CSV \\

ACO
& Motor target and measured states
& Avatar-side PC
& Avatar-side controller
& 10 Hz
& CSV \\

OCS
& Gaze vectors and image-plane gaze points
& Operator-side PC
& HMD eye tracking
& 10 Hz
& CSV \\

OCS
& Head orientation and head/wrist poses
& Operator-side PC
& HMD tracking
& 10 Hz
& CSV \\

OCS
& Facial blendshape weights
& Operator-side PC
& HMD face tracking
& 10 Hz
& CSV \\

OCS
& Arm joint targets
& Operator-side PC
& IK solver
& 10 Hz
& CSV \\

OCS
& Generated motor commands
& Operator-side PC
& Teleoperation controllers
& 10 Hz
& CSV \\

\hline
\end{tabular}
\end{table*}

\subsection{Android Avatar and Teleoperation System}

The framework comprises an operator-side teleoperation system, an avatar-side system deployed in the remote environment, and a network connecting the two sides.
Video and audio captured on the avatar side are transmitted to the operator and presented through a head-mounted display (HMD) and headphones.
Tracking information acquired by the HMD is converted into motion commands and transmitted to the avatar-side system.

We used the whole-body android avatar ``Yui'' as the embodied interface.
The hardware configuration, teleoperation interface, and previous real-world deployments of Yui are described in our prior work~\cite{Shinkawa2026Yui}.
Yui has 55 degrees of freedom and is equipped with actuated eyes, a facial-expression mechanism, a head, arms, and hands, a waist, and a mobile base.

Each eye contains a wide-angle camera with a 5-megapixel CMOS image sensor (OV5640, OMNIVISION) and a horizontal field of view of 200$^\circ$, while images were captured at a resolution of $1920 \times 1080$ pixels in our system.
Although both eye-camera streams were available, only the right-eye stream was used as the representative eye-camera image in this study. 
This avoided transient binocular inconsistencies caused by delays in avatar eye motion, thereby reducing discomfort during prolonged operation. 
The right-eye images were presented to both eyes of the HMD and used for data recording and gaze-point mapping.

A speakerphone (Jabra Speak2 55, GN Store Nord) mounted on the mobile base was used for live bidirectional voice communication.

An HMD (Meta Quest Pro, Meta) was used to acquire the operator's head, eye, face, and hand tracking data, which were then converted into avatar commands.
Head and eye orientations were converted into target values for the avatar's neck and eyes.
The HMD~\cite{Meta} provided 63 facial blendshape weights, which were converted into commands for the facial-expression mechanism.
For arm and waist motion, the operator's head and wrist poses acquired through HMD tracking were converted into target joint values using inverse kinematics (IK).
The avatar's eyes, facial expressions, neck, arms, and waist yaw axis were teleoperated.

A WebRTC-based communication system transmitted video, bidirectional audio, and control commands between the operator and avatar sides.
Recording functions were added to the teleoperation system to store ACOs and OCSs for subsequent analysis.






\subsection{Recorded Modalities and Data Logging}

Table~\ref{tab:recorded_modalities} summarizes the recorded data:
environmental sensory data, received control commands, and avatar motor states were classified as ACOs, whereas operator tracking data and the control information generated from them were classified as OCSs.

\begin{figure}[t]
    \centering
    \includegraphics[width=1.0\linewidth]{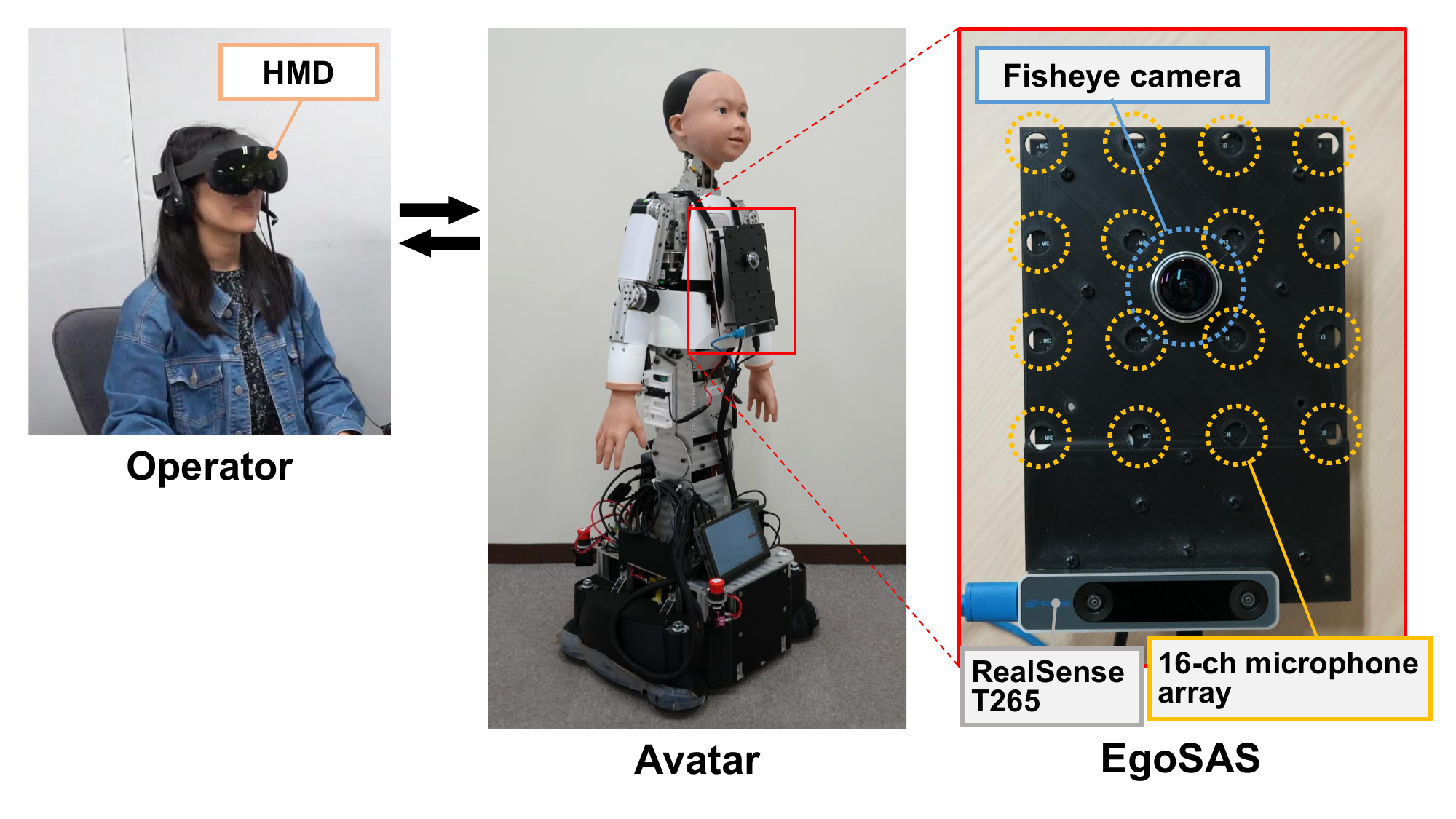}
    \caption{EgoSAS mounted on the android avatar. The Intel RealSense camera, used for visual-inertial odometry, was not used for data collection in this study.}
    \label{fig:yui_egosas}
\end{figure}

An Ego-centric Situational Awareness Sensor (EgoSAS)~\cite{Xu2025EgoSAS} was mounted on the avatar, as shown in Fig.~\ref{fig:yui_egosas}.
The EgoSAS used in this study captured fisheye images using a color CMOS camera board (DFM37UX273-ML, The Imaging Source) at a resolution of $540 \times 540$ pixels and 30~fps.
It also recorded 16-channel audio at 44.1~kHz using a microphone array (UMA-16 v2, miniDSP).
The multichannel audio can also be used to derive an acoustic map aligned with the EgoSAS image using the EgoSAS processing described in~\cite{Xu2025EgoSAS}, although this derived representation was not used in the analyses reported here.
Both streams were stored on a dedicated EgoSAS PC running Linux.
To reduce the risk of data loss during long-duration data collection, the recorded data were saved every 10 minutes.



The operator-side PC recorded the eye-camera images, gaze vectors, image-plane gaze points, head orientation, facial blendshape weights, head and wrist poses, arm joint targets, and generated motor commands.
The 3D gaze vector obtained from the HMD was converted in real time into an image-plane gaze point in the eye-camera image.
Its angular deviation from the forward direction was mapped to a pixel displacement from the image center using the camera's field of view and image resolution, with its horizontal and vertical components determining the displacement direction.
The mapping was qualitatively verified by directing gaze toward predefined locations in the displayed image.
The eye-camera images were recorded at 10~fps, and the numerical streams on the operator-side PC were logged at 10~Hz.

The avatar-side PC recorded the received control commands together with the target and measured states of the avatar motors at 10~Hz.
These recording rates were selected to reduce the computational load.



\subsection{Temporal and Spatial Data Association}

The recorded streams were associated across devices using timestamps. 
Visual ACOs and gaze-related OCSs were additionally associated through image-based spatial mapping.

\subsubsection{Temporal Association}
\label{sec:temporal_association}

Each data stream was assigned a Unix timestamp obtained from the system clock of the corresponding recording device.
No dedicated clock synchronization was performed between the recording devices.
Temporal association was performed only between the EgoSAS and operator-side recordings, as the avatar-side recordings served only as supplementary logs of the avatar state.

To estimate the temporal offset between the EgoSAS and operator-side recordings, a shared instantaneous visual event visible in both the EgoSAS and avatar eye camera was recorded during data collection.
During offline processing, the timestamps of the frames capturing this event were compared to estimate the offset.
The estimated offset was used to assess whether timestamp correction was required.
The overlapping interval between the two recordings was then extracted.
Within this interval, the eye-camera and EgoSAS image sequences were reconstructed at effective frame rates calculated from the timestamp span and frame count of each sequence.



\subsubsection{Spatial Association of Visual ACOs and Gaze OCSs}
\label{sec:data_processing}

As a concrete implementation of the spatial association of gaze-related OCSs with visual ACOs, the operator's image-plane gaze point was mapped from the eye-camera image onto the corresponding EgoSAS image.
Although this mapping introduces additional spatial uncertainty, it enables gaze to be associated with a stable, wide-angle view of the interaction scene that is unaffected by the avatar's eye movements and eyelid occlusions.
The EgoSAS fisheye images were converted offline into a latitude--longitude representation using OpenCV-based remapping followed by a custom spherical-coordinate transformation.
The camera matrix used the image center and an empirically selected focal length; the distortion coefficients were set to zero.

The cross-camera mapping was performed offline once every three EgoSAS frames, corresponding to approximately 10~Hz.
SuperPoint~\cite{detone2018superpoint} was used to extract local features from the eye-camera images and unwrapped EgoSAS images, and LightGlue~\cite{lindenberger2023lightglue} was used to establish feature correspondences.
When at least 20 feature correspondences were available, a homography was estimated using OpenCV's \texttt{findHomography} function with RANSAC.
The homography was considered valid if RANSAC yielded at least 12 inliers and an inlier ratio of at least 25\%.
Each valid homography was used to map the image-plane gaze point onto the corresponding unwrapped EgoSAS image.
When a valid homography could not be obtained, the most recent valid homography was reused for up to 1~s (30 EgoSAS frames); frames beyond this interval were excluded from subsequent gaze-related processing.

\begin{figure}[!t]
    \centering
    \begin{tabular}{cc}
        \begin{minipage}{0.52\columnwidth}
            \centering
            \begin{subfigure}[t]{\columnwidth}
                \centering
                \includegraphics[height=2.9cm]{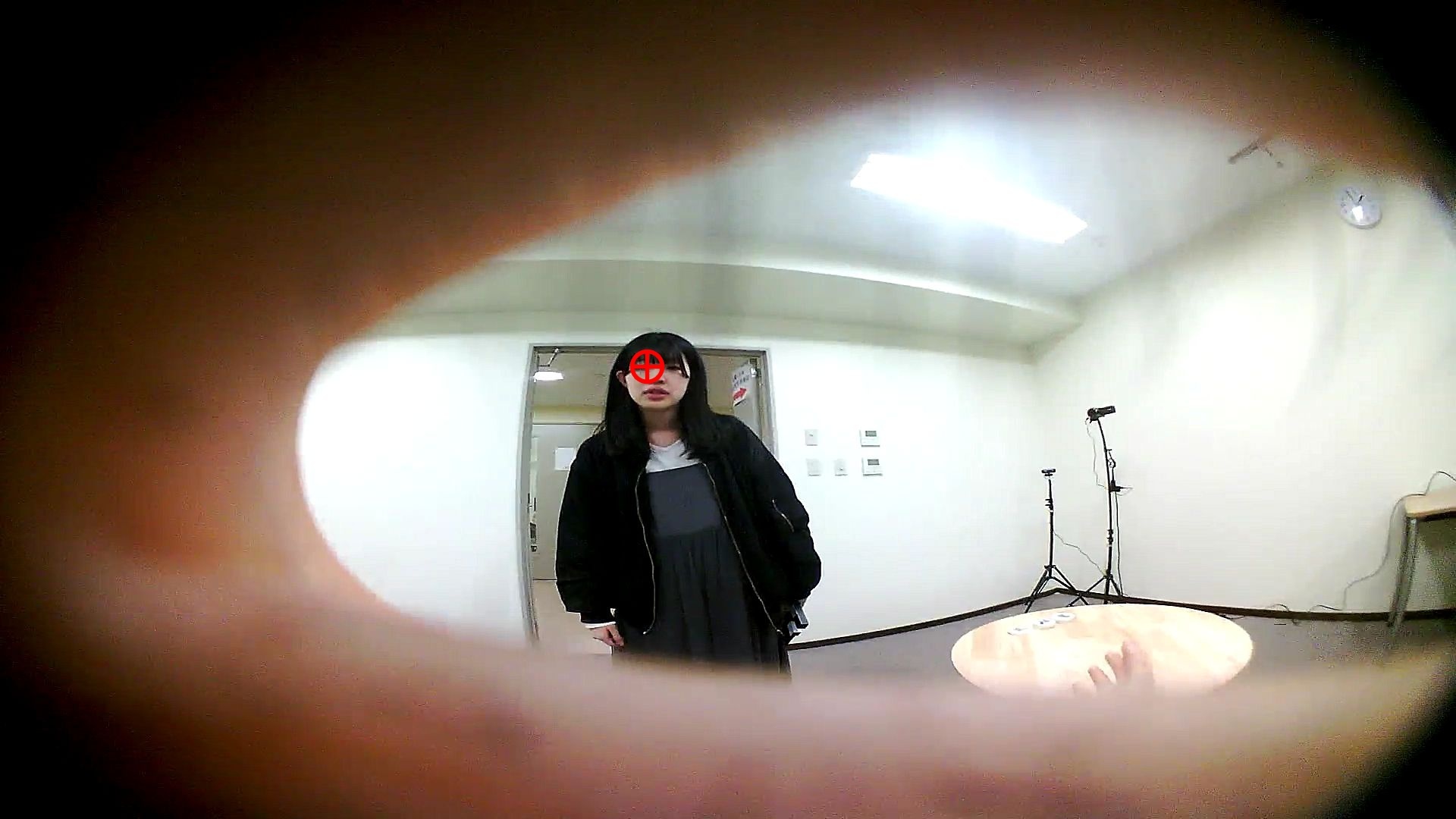}
                \subcaption{Eye-camera image}
                \label{fig:eyecam_gaze}
            \end{subfigure}
        \end{minipage}
        &
        \begin{minipage}{0.41\columnwidth}
            \centering
            \begin{subfigure}[t]{\columnwidth}
                \centering
                \includegraphics[height=2.9cm]{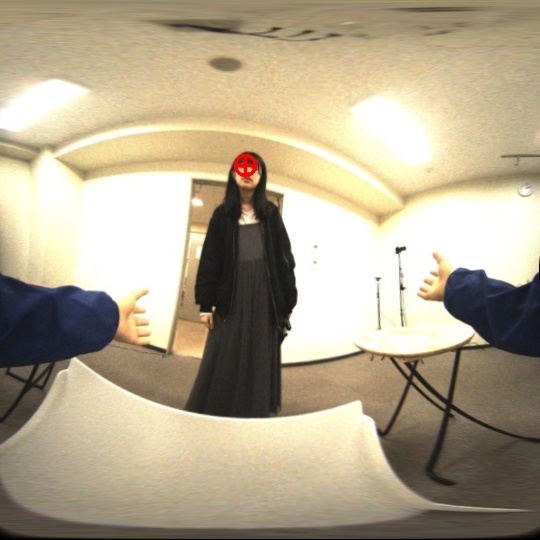}
                \subcaption{EgoSAS image}
                \label{fig:fisheye_gaze}
            \end{subfigure}
        \end{minipage}
    \end{tabular}
    \caption{Illustrative example of mapping an image-plane gaze point from (a) the eye-camera image to (b) the unwrapped EgoSAS image. The images are for demonstration only and were not included in the real-world deployments described in Section~\ref{sec:data_collection}. The person shown is one of the authors.}
    \label{fig:img_gaze_point}
\end{figure}

Fig.~\ref{fig:img_gaze_point} shows an illustrative example of the gaze-point mapping.
This spatial association enables gaze-related OCSs to be examined in relation to image regions and people or objects detected in the visual ACOs.




\section{Real-World Data Collection}
\label{sec:data_collection}

Using the framework described in Section~\ref{sec:framework}, we collected android-avatar-mediated social interaction data in two distinct real-world settings: a university open-lab event and a corporate office lobby.
Both deployments used the same modalities and logging procedures but differed in their interaction contexts, participants, and operators.

\subsection{Common Procedure and Ethical Considerations}

Participation in both deployments was limited to adults aged 18 years or older.
The participants were informed of the study through notices and instructions before participation.
In accordance with the approved protocol, entering the designated study area was considered as consent to participate.
The participants communicated freely with the operator through the avatar, with no fixed time limit, and could end the interaction at any time.

The protocol for both deployments was approved by the Ethics Committee of the University of Electro-Communications (No.~H25113), and data collection followed the Declaration of Helsinki.

In both settings, two external video cameras were installed behind and to the front-right of the avatar to document the collection conditions from third-person viewpoints.
These recordings were used only to check the collection conditions and were not included in the analyses reported in this study.

For the offline temporal association described in Section~\ref{sec:temporal_association}, a shared visual reference event visible to both the EgoSAS and eye camera was recorded in each deployment.
A clapperboard closure was used in the university setting, whereas the instant of hand contact during a clap was used in the corporate office setting.

\subsection{Deployment Settings}

\subsubsection{University Open-Lab Setting}

The first deployment was conducted during an open-lab event organized by the University of Electro-Communications.
The avatar was installed in a room open to visitors, who were invited to freely ask questions to laboratory students through the avatar.
Data collection continued for approximately four hours, during which 34 visitors (26 groups) participated.
Four laboratory students, including two of the authors, operated the avatar in shifts, and the operator-side recordings were stored separately for each operator.
The layout and appearance of this environment are shown in
Fig.~\ref{fig:deployment_settings}(a) and (b).

After the interaction, the participants were invited to complete an optional four-point questionnaire asking whether guidance provided through the android avatar could substitute for guidance provided by a physically present person.


\subsubsection{Corporate Office Setting}

\begin{figure}[!t]
    \centering

    \begin{subfigure}[t]{0.30\columnwidth}
        \centering
        \includegraphics[height=3.7cm]{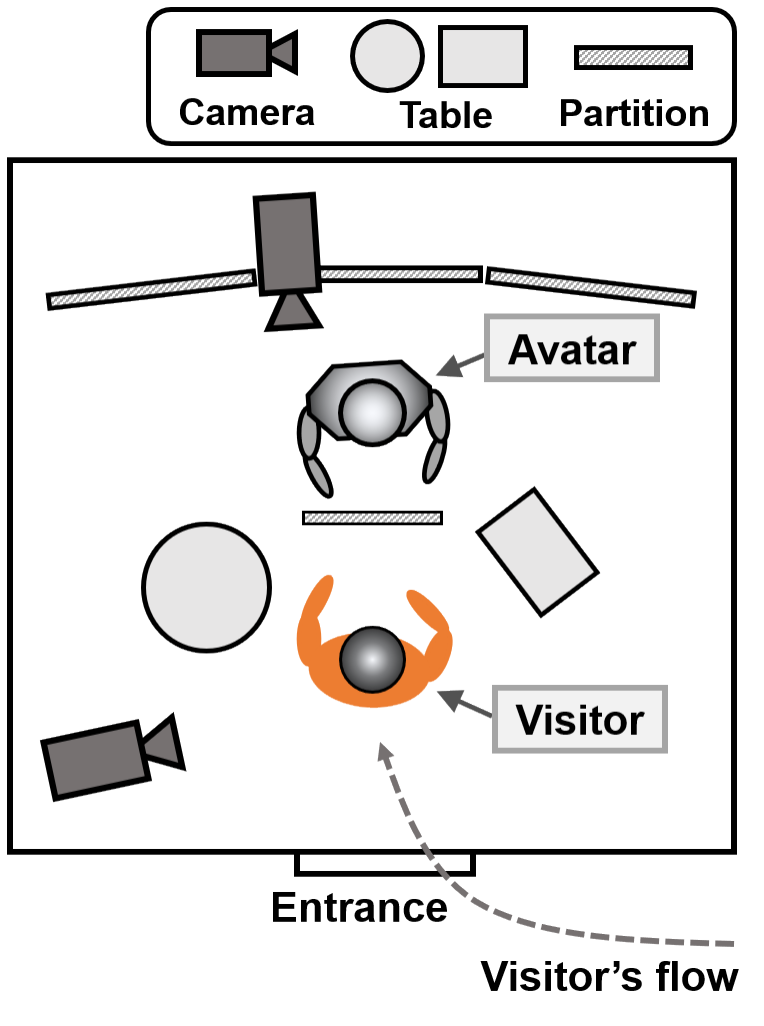}
        \subcaption{Open-lab layout}
        \label{fig:univ_layout}
    \end{subfigure}
    \hfill
    \begin{subfigure}[t]{0.65\columnwidth}
        \centering
        \includegraphics[height=3.7cm]{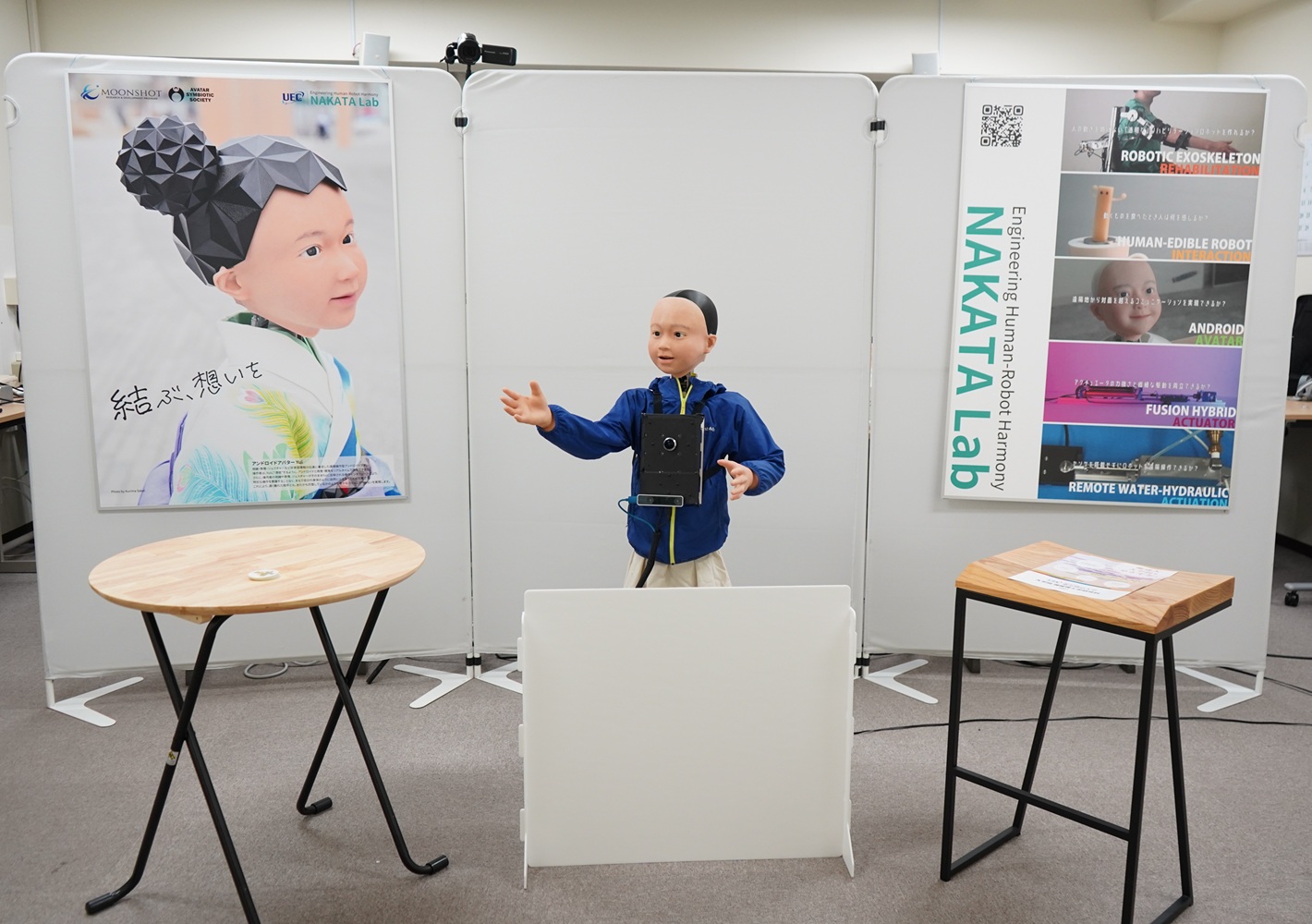}
        \subcaption{Open-lab photograph}
        \label{fig:univ_photo}
    \end{subfigure}

    \vspace{0.4em}

    \begin{subfigure}[t]{0.30\columnwidth}
        \centering
        \includegraphics[height=3.7cm]{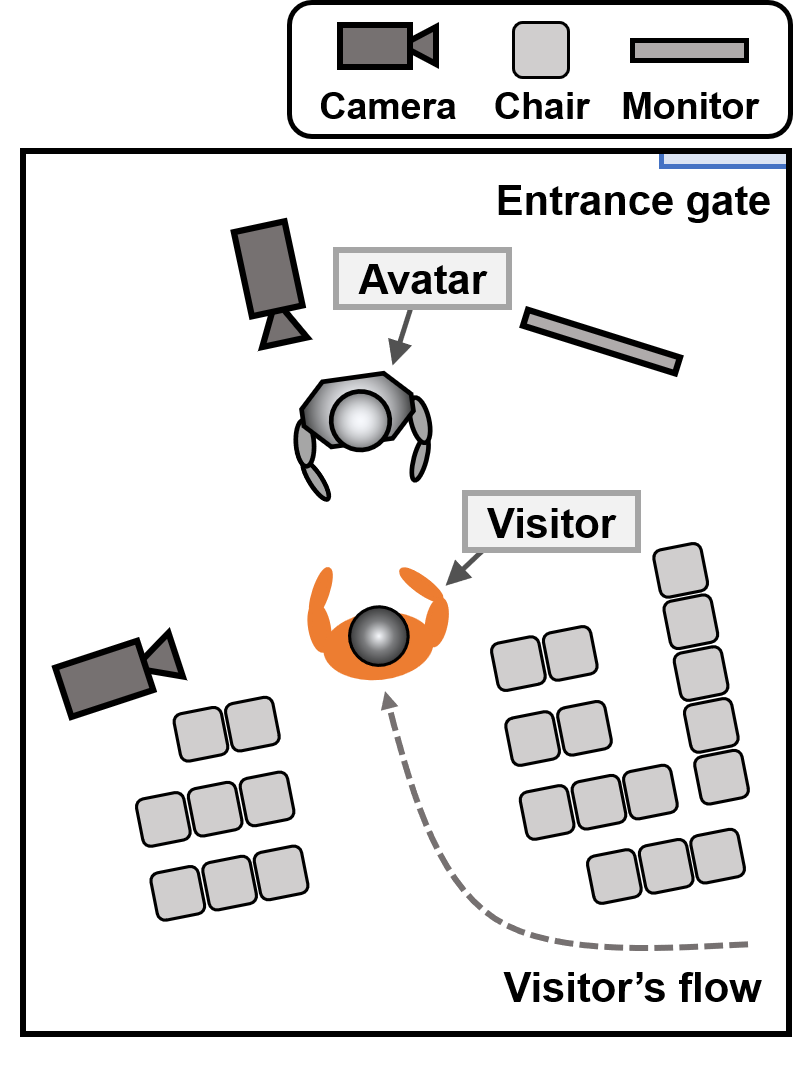}
        \subcaption{Office layout}
        \label{fig:office_layout}
    \end{subfigure}
    \hfill
    \begin{subfigure}[t]{0.65\columnwidth}
        \centering
        \includegraphics[height=3.7cm]{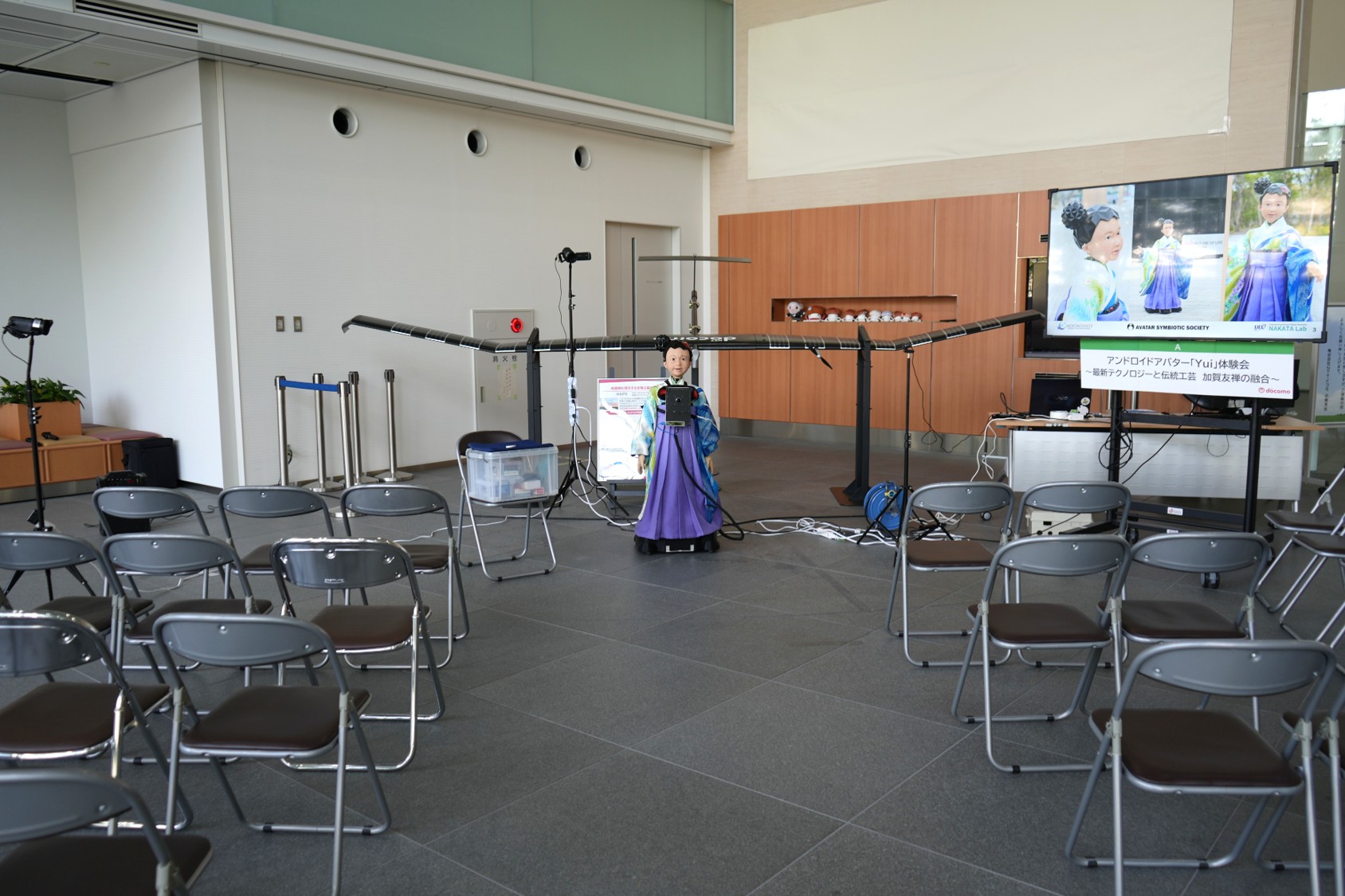}
        \subcaption{Office photograph}
        \label{fig:office_photo}
    \end{subfigure}

    \caption{Data-collection setups in the two real-world deployment
    settings: (a, b) university open-lab environment and
    (c, d) corporate office environment.}
    \label{fig:deployment_settings}
\end{figure}

The second deployment was conducted in the lobby of the NTT DOCOMO Hokuriku Regional Office.
Company employees were informed in advance about the android-avatar interaction event and were invited to participate.
Data collection continued for approximately thirty minutes, during which 16 visitors (7 groups) participated.
The avatar was operated by one of the authors throughout the deployment.
The layout and appearance of this environment are shown in
Fig.~\ref{fig:deployment_settings}(c) and (d).

\subsection{Collected Data Summary}

Table~\ref{tab:data_summary} summarizes the number of operators and visitors, the data collection duration, and other details for each setting.
The total durations reported in the table include operator-change periods and other intervals without overlapping recordings.
Across both settings, data collection lasted 268 minutes (approximately 4.5 hours) and involved four operators and 50 visitors comprising 33 groups (19 individuals, 12 pairs, one group of three, and one group of four).

Using the procedure described in
Section~\ref{sec:temporal_association}, the observed timestamp difference between the EgoSAS and operator-side recordings was less than 0.1~s at the reference event in each deployment.
Because this difference did not materially affect the interaction-scale analyses reported in this study, no timestamp correction was applied.

The optional questionnaire in the open-lab setting was completed by 24 visitors aged from their late teens to their fifties.
When asked whether avatar-based guidance could substitute for guidance provided by a physically present person, 11 responded ``to a great extent'' and 13 answered ``to some extent''; none selected a negative option.

\begin{table}[t]
    \centering
    \caption{Summary of the two real-world data collections.}
    \label{tab:data_summary}
    \footnotesize
    \setlength{\tabcolsep}{2.4pt}
    \begin{tabular}{lrrrrr}
        \hline
        Setting
        & Operators\hspace{0.4em}
        & Visitors
        & \shortstack{Groups}
        & \shortstack{Total duration\\(minutes)}
        & \shortstack{Valid interaction\\duration (minutes)} \\
        \hline
        Open lab & 4\hspace{0.4em} & 34 & 26 & 240 & 135 \\
        Office   & 1\hspace{0.4em} & 16 &  7 &  28 &  21 \\
        \hdashline[2pt/1pt]
        Total    & 4$^{\dagger}$ & 50 & 33 & 268 & 156 \\
        \hline
    \end{tabular}
    
    \vspace{1mm}
    {\scriptsize\selectfont
    \raggedright
    $^{\dagger}$The operator in the office setting overlapped with those in the open-lab setting.
    \par
    }
\end{table}

\textbf{Interaction Segment Extraction:}
We first extracted periods with overlapping EgoSAS and operator-side recordings.
Within these periods, we identified continuous intervals with a visitor present in front of the avatar and valid operator-side HMD tracking.
Person detection was performed every 0.5~s on the EgoSAS images using YOLOv5~\cite{yolov5}.
Detections with confidence below 0.3, bounding-box centers outside the central 50\% of the image width, or bounding-box heights below 25\% of the image height were discarded to reduce detections of bystanders, distant visitors, and the avatar itself.
A visitor was considered present when the median number of retained detections over five frames was at least one.

For each HMD head, eye, and face tracker, intervals with unchanged values for at least three consecutive samples were considered tracking failures.
Finally, continuous intervals of at least 30~s containing a visitor and no tracking failures were extracted as interaction segments.

This procedure yielded 46 interaction segments in total, including 38 from the open-lab setting and 8 from the office setting. 
Their total duration was 156 minutes (2.6 hours), comprising 135 minutes from the open-lab setting and 21 minutes from the office setting.
Among the periods with overlapping EgoSAS and operator-side recordings, 43 minutes were excluded, primarily due to visitor absence alone (74.6\%), followed by tracking failures alone (23.0\%).

\section{Exploratory Analysis of Operator-Centric Social Cues}
\label{sec:analysis}

This section has two purposes: to assess the scene-level consistency of the cross-camera gaze-point mapping and to illustrate how linked ACOs and OCSs can support temporal and contextual analyses of operator social cues.
The primary analysis relates face-directed gaze to subsequent speech onset using spatially and temporally associated visual ACOs and gaze OCSs; complementary analyses relate smile-related facial signals and operator social cues to speech events and states.
These analyses are exploratory and are not intended to establish generalizable behavioral findings.

\subsection{Scene-Level Assessment of Cross-Camera Gaze-Point Mapping}
\label{sec:gaze_mapping_assessment}
We randomly sampled 100 image pairs from periods when both cameras were recording, accounting for differences in deployment durations: 80 from the open-lab setting and 20 from the office setting.
Each pair comprised an eye-camera image with the original gaze point and an EgoSAS image with the mapped point.
Two authors not involved in image creation independently classified whether the points indicated the same person, object, body part, or scene region as consistent, inconsistent, or indeterminate.
Disagreements were conservatively classified as indeterminate, and inter-annotator agreement was assessed using Cohen's $\kappa$.

The results showed that of the 100 image pairs, 66 were classified as consistent, 11 as inconsistent, and 23 as indeterminate. The consistency rate was 85.7\% among judgeable samples. 
The two annotators achieved an agreement rate of 77.0\% and a Cohen's $\kappa$ of 0.42.

\subsection{Analysis Data and Event Labeling}

From the 38 open-lab interaction segments, we selected one-to-one interactions to avoid differences in interaction dynamics between one-to-one and group conversations.
We limited segment duration to 2--4 min to reduce variation in segment length and retained only the longest segment per visitor to avoid repeated inclusion.
All seven segments meeting these criteria were analyzed, with each of the four operators represented in at least one segment.

\subsubsection{Speech Event Labeling}

WhisperX~\cite{bain2022whisperx} was first used to obtain timestamped transcriptions and initial speaker diarization for the selected segments.
Based on these outputs, one of the authors manually annotated speech at a temporal resolution of 0.1~s.
Backchannel utterances and fillers were included as speech, whereas nonverbal vocalizations such as laughter were excluded.
Each time point was then assigned one of four speech states: operator speaking, visitor speaking, both speaking, and no speech.

\subsubsection{Operator-Centric Social Cue Labeling}

Gaze coordinates and HMD Face Tracker data were sampled at 0.1-s intervals.
For gaze labeling, the operator's image-plane gaze point was mapped onto the EgoSAS image using the procedure described in Section~\ref{sec:data_processing}.
Faces were detected in the EgoSAS images using YOLOv11n face detection~\cite{yolov11nface}.
A time point was labeled as face-directed gaze if a detected face bounding-box center fell within a circle centered at the gaze point with a radius of 10\% of the image width.

For facial-expression labeling, a smile-related value was computed as the mean of four HMD blendshapes:
\texttt{CHEEK\_RAISER\_L},
\texttt{CHEEK\_RAISER\_R},
\texttt{LIP\_CORNER\_ PULLER\_L}, and
\texttt{LIP\_CORNER\_PULLER\_R}~\cite{Meta}.
To account for variability in blendshape magnitude, smiling and smile-onset events were defined using percentile-based thresholds within each interaction segment.
The top 5\% of time points ranked by this value were labeled as smiling.
The increase from the preceding sampled time point was also calculated, and the top 5\% of increases were labeled as smile-onset events.
When consecutive time points met this criterion, only the first was retained as an onset event.

\subsection{Exploratory Social Cue--Speech Relationships}

All analyses were restricted to the interval between the onset of the first utterance and the end of the final utterance in each selected segment.
For each analysis, segments whose numbers of analyzed frames exceeded the third quartile by more than three times the interquartile range were excluded to reduce the influence of any single segment.

\subsubsection{Face-Directed Gaze and Subsequent Speech Onset}

This analysis examined whether speech-onset rates differed following periods with and without face-directed gaze during no-speech intervals.
Intervals lasting at least 1~s were eligible.
To avoid sampling adjacent frames, one analysis frame was sampled after the first 1~s of continuous no-speech state and every 0.5~s thereafter.
One segment whose number of eligible frames exceeded the third quartile by more than three times the interquartile range was excluded.
Face-directed gaze was classified as present if detected for more than 0.5~s during the preceding 1-s interval and absent if detected for less than 0.5~s; frames with exactly 0.5~s were excluded.
We then calculated the observed rates at which the operator or visitor began speaking within the subsequent 1 and 2~s.

\begin{figure}[!t]
    \centering
    \begin{tabular}{cc}
        \begin{minipage}{0.47\columnwidth}
            \centering
            \begin{subfigure}[t]{\columnwidth}
                \centering
                \includegraphics[width=\columnwidth]
                {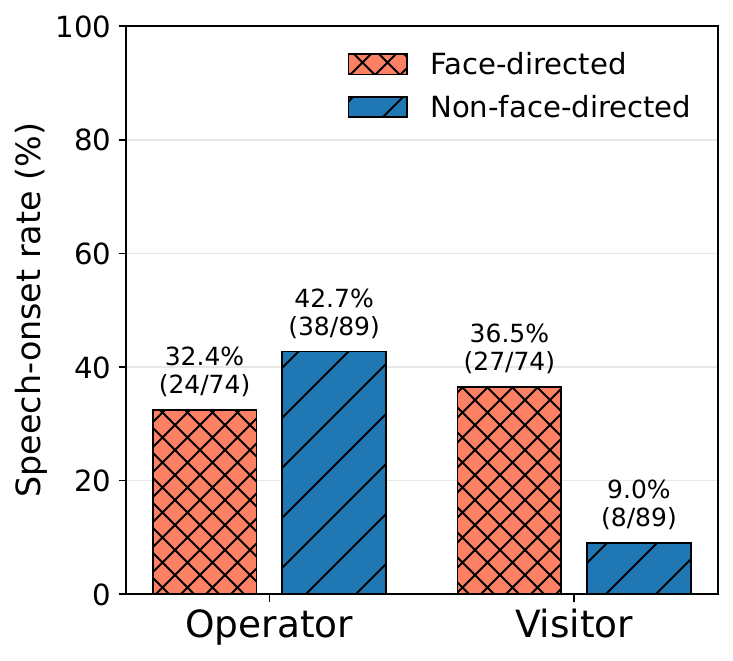}
                \subcaption{Speech onset within 1~s.}
                \label{fig:gaze_speech_onset_1sec}
            \end{subfigure}
        \end{minipage}
        &
        \begin{minipage}{0.47\columnwidth}
            \centering
            \begin{subfigure}[t]{\columnwidth}
                \centering
                \includegraphics[width=\columnwidth]
                {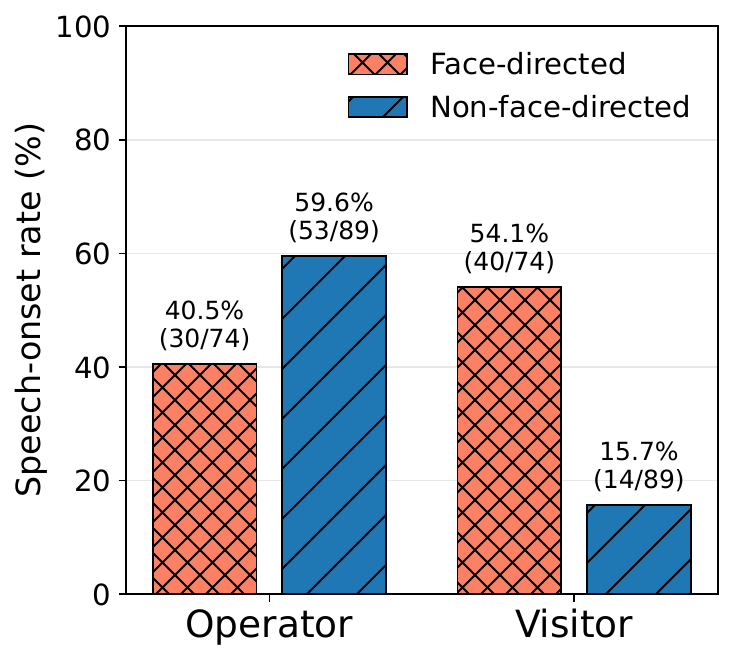}
                \subcaption{Speech onset within 2~s.}
                \label{fig:gaze_speech_onset_2sec}
            \end{subfigure}
        \end{minipage}
    \end{tabular}
    \caption{Observed operator and visitor speech-onset rates within (a) 1~s and (b) 2~s following eligible analysis frames with and without preceding face-directed gaze. Labels show percentages (count/total).}
    \label{fig:gaze_speech_onset}
\end{figure}

The results are shown in Fig.~\ref{fig:gaze_speech_onset}.
Within the analyzed segments, operator speech onset was observed more frequently in the absence of face-directed gaze, whereas visitor speech onset was observed more frequently following face-directed gaze, for both the 1- and 2-s windows.

\subsubsection{Smile Onset and Subsequent Speech Onset}
As a complementary analysis of facial-expression OCSs, we examined speech onset following smile-onset frames.
For each speaker, frames where that speaker was already speaking were excluded.
For each remaining smile-onset frame, we determined whether the operator or visitor began speaking within 1 or 2~s. 
The same rates were also calculated across all eligible analysis frames, regardless of smile onset, to provide a descriptive baseline.

\begin{figure}[!t]
    \centering
    \begin{tabular}{cc}
        \begin{minipage}{0.47\columnwidth}
            \centering
            \begin{subfigure}[t]{\columnwidth}
                \centering
                \includegraphics[width=\columnwidth]
                {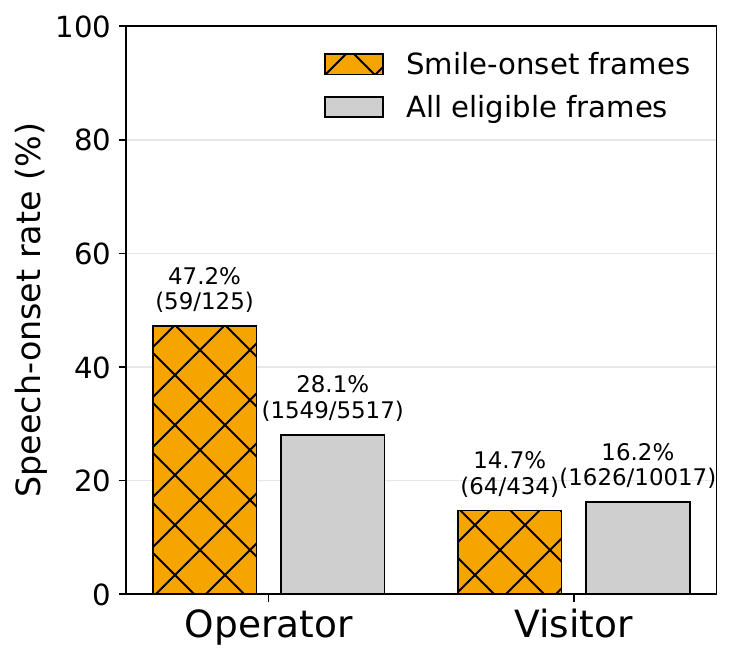}
                \subcaption{Speech onset within 1~s.}
                \label{fig:smile_speech_onset_1sec}
            \end{subfigure}
        \end{minipage}
        &
        \begin{minipage}{0.47\columnwidth}
            \centering
            \begin{subfigure}[t]{\columnwidth}
                \centering
                \includegraphics[width=\columnwidth]
                {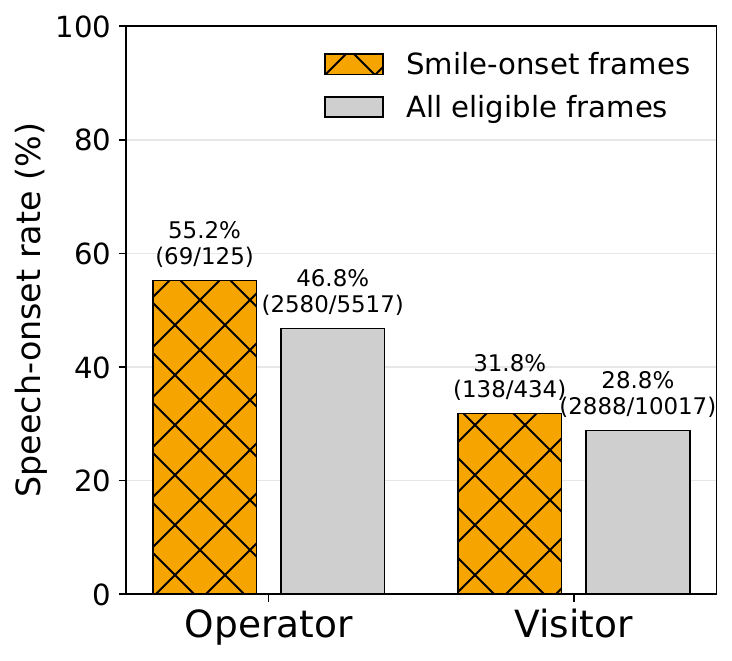}
                \subcaption{Speech onset within 2~s.}
                \label{fig:smile_speech_onset_2sec}
            \end{subfigure}
        \end{minipage}
    \end{tabular}
    \caption{Observed operator and visitor speech-onset rates within (a) 1~s and (b) 2~s at smile-onset frames and across all eligible frames. Labels show percentages (count/total).}
    \label{fig:smile_speech_onset}
\end{figure}

The results are shown in Fig.~\ref{fig:smile_speech_onset}.
Within the analyzed segments, operator speech onset was observed more frequently at smile-onset frames than across all eligible frames within both 1 and 2~s, although the difference was smaller for the 2-s window. 
For visitor speech onset, only small differences were observed between smile-onset and all eligible frames for both time windows.

\subsubsection{Operator Social Cues Across Speech States}

Finally, we examined how the occurrence of operator social cues varied with the current speaker.
For frames labeled as operator or visitor speaking, we calculated the observed proportions in which the operator exhibited face-directed gaze toward the visitor or was labeled as smiling.

\begin{figure}[t]
    \centering
    \includegraphics[width=0.65\columnwidth]
    {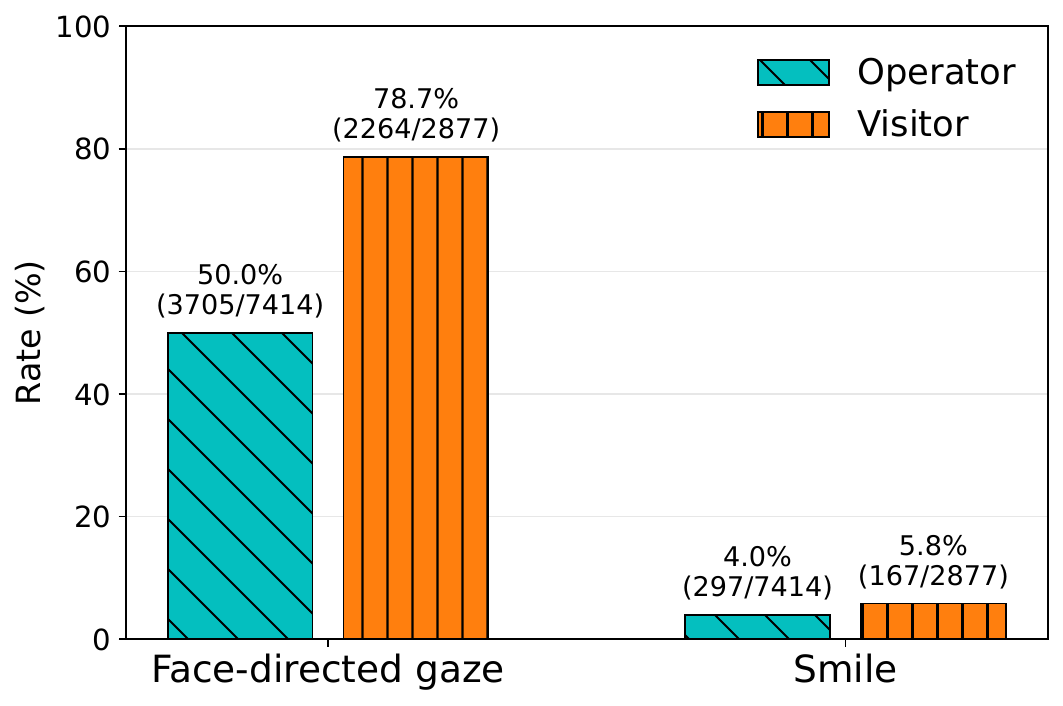}
    \caption{Observed occurrence rates of face-directed gaze and smiling during operator and visitor speech.
Labels show percentages (count/total).}
    \label{fig:speaker_state_social_cues}
\end{figure}

As shown in Fig.~\ref{fig:speaker_state_social_cues}, face-directed gaze was observed more frequently during visitor speech than during operator speech.
Smiling was also observed more frequently during visitor speech.


\section{Discussion}
\label{sec:discussion}

A central contribution of this study is the development of a multimodal data-collection framework that records ACOs and OCSs as complementary descriptions of the same embodied social interaction.
RoboTrio2~\cite{EliseiHB24} and our framework both associate robot-side observations with teleoperator behavior but differ in recorded signals and their linkage to the interaction scene, as summarized in Table~\ref{tab:related_frameworks}.
The contribution therefore lies in integrating scene-mapped gaze, operator facial, head, and upper-body signals, control commands, and avatar motor states within a single data-collection architecture using a whole-body android embodiment, rather than in the performance of any individual sensing or analysis method.
Although the exploratory analyses in this study focus on gaze, facial expressions, and speech, the recorded upper-body motion signals also provide a basis for future analyses of embodied social behaviors such as gestures.

The framework was operated for 4.5 hours across two real-world social settings that differed in their environments, operators, visitors, and interaction purposes.
These deployments show that the same recording framework can be applied beyond a single controlled laboratory protocol.
The proposed framework also offers a practical advantage for multimodal data collection.
Unlike human--human interaction studies that may require multiple wearable sensors, it captures operator behavioral signals using only an HMD, while the interlocutor interacts with the android avatar rather than a person wearing multiple measurement devices.

The cross-camera gaze mapping provided preliminary evidence of scene-level consistency among judgeable samples, supporting its use in exploratory analyses of face-directed gaze.
However, analyses requiring finer spatial precision would need further calibration and validation.

The face-directed-gaze analyses showed different descriptive patterns across conversational roles.
During no-speech intervals, visitor speech onset was more frequent following face-directed gaze, whereas operator speech onset was more frequent in its absence.
Face-directed gaze was also more frequent during visitor speech than operator speech, qualitatively consistent with prior observations that listeners tend to look toward the person they are attending to~\cite{Admoni2017}.

For smile-related signals, a descriptive tendency was observed for operator speech onset to occur more frequently at smile-onset frames than across all eligible frames.
This short-lag pattern is broadly consistent with conversation-analytic observations that smiles can precede the opening of a spoken turn~\cite{Kaukomaa2013}.
However, because smile onsets were identified using percentile thresholds rather than validated labels, their interpretation as conversational signals remains uncertain.

Taken together, these results show that this study provides an integrated data-collection architecture for examining relationships between avatar-observable contexts and operator-demonstrated responses within the same embodied interaction, rather than establishing general behavioral regularities.







\subsection{Limitations}
The dataset collected in this study covered only two real-world settings, with limited diversity in operators, visitors, and interaction scenarios.
The exploratory analyses were further limited to seven one-to-one interaction segments, restricting the generalizability of the findings and leaving interaction dynamics in one-to-many settings unexplored.
Additionally, public release of the collected recordings and annotations is constrained by privacy concerns, while privacy-preserving approaches to data sharing and framework reuse will be explored in future work.

Temporal and spatial association also had limitations. 
Temporal association relied on device timestamps and a shared visual reference event in each deployment. 
Although the timestamp difference at this event was less than 0.1~s in both deployments, this single-point comparison could not characterize possible time-varying transmission or processing delays during recording. 
Spatial association between gaze OCSs and visual ACOs was evaluated only for scene-level consistency, and further validation is needed to quantify gaze-mapping errors.

Finally, the exploratory analyses had limitations in annotation reliability and contextual consideration.
Speech states were annotated by one author without assessing inter-rater reliability, and smile onsets were identified using percentile thresholds without validation against manually annotated labels.
Furthermore, because conversational content, participant relationships, operator-specific behavior, and other contextual factors were not modeled, the observed patterns should not be generalized beyond the analyzed segments.

\subsection{Future Work}

Future work will expand data collection across more diverse environments, operators, visitors, and interaction scenarios, while improving temporal and spatial association through periodic synchronization events and clock drift assessment, as well as annotation reliability.
With larger and more diverse datasets, we will compare ACO-only and OCS-assisted social-interaction world models and evaluate their ability to predict interaction events and generate socially appropriate behavior from ACOs and interaction history during autonomous avatar
operation.

\section{Conclusion}
\label{sec:conclusion}
In this study, we developed a multimodal data-collection framework that records and associates ACOs and OCSs during the same avatar-mediated social interaction.
The framework was deployed in two real-world environments, yielding 4.5 hours of multimodal recordings, including 2.6 hours of valid social-interaction segments.
Exploratory analyses demonstrated how the linked recordings can support temporal and contextual analyses of operator social cues.
This framework provides a foundation for future social-interaction world models that relate avatar-observable interaction contexts to human-demonstrated social behavior and support autonomous socially interactive avatars.
\bibliographystyle{IEEEtran}
\bibliography{IEEEexample.bib}

\end{document}